\documentclass[twoside,11pt]{article}

\usepackage{jmlr2e}

\usepackage{hyperref}       % hyperlinks
\usepackage{url}            % simple URL typesetting
\usepackage{amsfonts}       % blackboard math symbols
\usepackage{amsmath}
\usepackage{graphicx} % Allows including images
\usepackage{subcaption}
\usepackage{amssymb}
\usepackage{algorithm}
\usepackage[noend]{algpseudocode}

\ShortHeadings{Identifying Generalization Properties in Neural Networks}{Huan, Nitish, Caiming, and Richard}
\firstpageno{1}

\begin{document}

%\title{Hessian, Sharpness Metric, and Generalization}
\title{Identifying Generalization Properties in Neural Networks}

\author{\name Huan Wang, \name Nitish Shirish Keskar,  \name Caiming Xiong, \name Richard Socher \\
\addr Salesforce Research\\
\email \{huan.wang, nkeskar, cxiong, rsocher\}@salesforce.com
       }

\editor{}

\maketitle

\begin{abstract}%   <- trailing '%' for backward compatibility of .sty file
While it has not yet been proven, empirical evidence suggests that model generalization is related to local properties of the optima which can be described via the Hessian.
We connect model generalization with the local property of a solution under the PAC-Bayes paradigm. In particular, we prove that model generalization ability is related to the Hessian, the higher-order ``smoothness" terms characterized by the Lipschitz constant of the Hessian, and the scales of the parameters. Guided by the proof, we propose a metric to score the generalization capability of the model, as well as an algorithm that optimizes the perturbed model accordingly. 
\end{abstract}

\begin{keywords}
generalization, PAC-Bayes, Hessian, perturbation.
\end{keywords}

\section{Introduction}

Deep models have proven to work well in applications such as computer vision \citep{Krizhevsky2012} \citep{HeZR014} \citep{Karpathy2014}, speech recognition \citep{Mohamed2012} \citep{Hinton12}, and natural language processing \citep{socher2013} \citep{Graves13} \citep{mccann2018}. Many deep models have millions of parameters, which is more than the number of training samples, but the models still generalize well \citep{huang2017densely}. 

On the other hand, classical learning theory suggests the model generalization capability is closely related to the ``complexity" of the hypothesis space. This seems to be a contradiction to the empirical observations that over-parameterized models generalize well on the test data. Indeed, even if the hypothesis space is complex, the final solution learned from a given training set may still be simple. An example is, suppose the hypothesis space is the union of linear classifiers and some complex function spaces. As a union set the hypothesis space is complex in the worst case, but for some training set the best solution may be a linear classifier. This suggests the generalization capability of the model is also related to the property of the solution.

\cite{Keskar16} and \cite{ChaudhariCSL16} empirically observe that the generalization ability of a model is related to the spectrum of the Hessian matrix $\nabla^2 L(w^\ast)$ evaluated at the solution, and large eigenvalues of the $\nabla^2 L(w^\ast)$ often leads to poor model generalization. Also, \citep{Keskar16}, \citep{ChaudhariCSL16} and \citep{novak2018sensitivity} introduce several different metrics to measure the ``sharpness" of the solution, and demonstrate the connection between the sharpness metric and the generalization empirically. \cite{Dinh2017} later points out that most of the Hessian-based sharpness measures are problematic and cannot be applied directly to explain generalization. In particular, they show that the geometry of the parameters in RELU-MLP can be modified drastically by re-parameterization.

Another line of work originates from the theorists. \citep{Langford2001} and more recently \citep{harvey2017} \citep{Neyshabur2017} \citep{Neyshabur2017a} use PAC-Bayes bound to analysis the generalization behavior of the deep models. Since the PAC-Bayes bound holds uniformly for all ``posteriors", it also holds for some particular ``posteriors", for example, the solution parameter perturbed with noise. This provides a natural way to incorporate the local property of the solution into the generalization analysis. In particular, \cite{Neyshabur2017} suggests to use the difference between the perturbed loss and the empirical loss as the sharpness metric. \cite{Dziugaite2017} tries to optimize the PAC-Bayes bound instead for a better model generalization. Still some fundamental questions remain unanswered. In particular we are interested in the following question:\\

\emph{How is model generalization related to local ``smoothness" of a solution?}\\

In this paper we try to answer the question from the PAC-Bayes perspective. Under mild assumptions on the Hessian of the loss function, we prove the generalization error of the model is related to this Hessian, the Lipschitz constant of the Hessian, the scales of the parameters, as well as the number of training samples. The analysis also gives rise to a new metric for generalization. Based on this, we can approximately select an optimal perturbation level to aid generalization which interestingly turns out to be related to Hessian as well. Inspired by this observation, we propose a perturbation based algorithm that makes use of the estimation of the Hessian to improve model generalization.

\begin{figure}
\hskip -0.3in
\begin{subfigure}{.6\textwidth}
  \centering
  \includegraphics[height=2.5in]{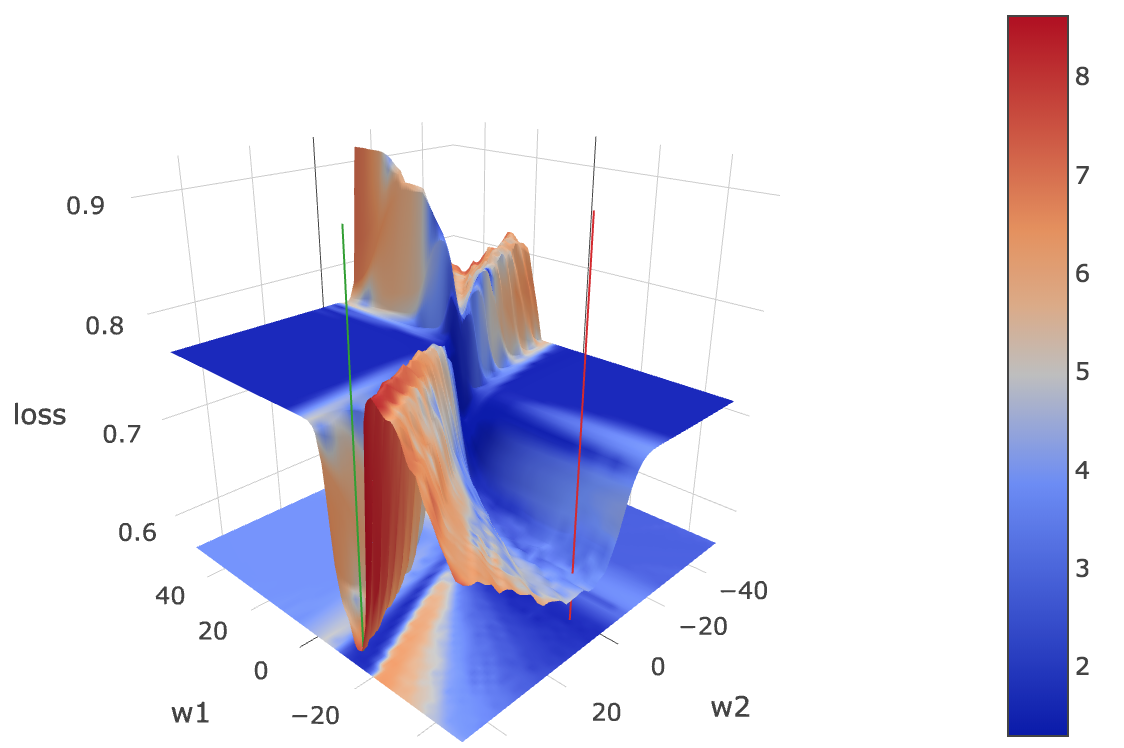}
\caption{Loss landscape. The color on the loss surface shows the pacGen scores. The color on the bottom plane shows an approximated generalization bound.}  \label{fig:toy_metrics}
\end{subfigure}
\begin{subfigure}{.6\textwidth}
  \centering
  \includegraphics[height=2.5in]{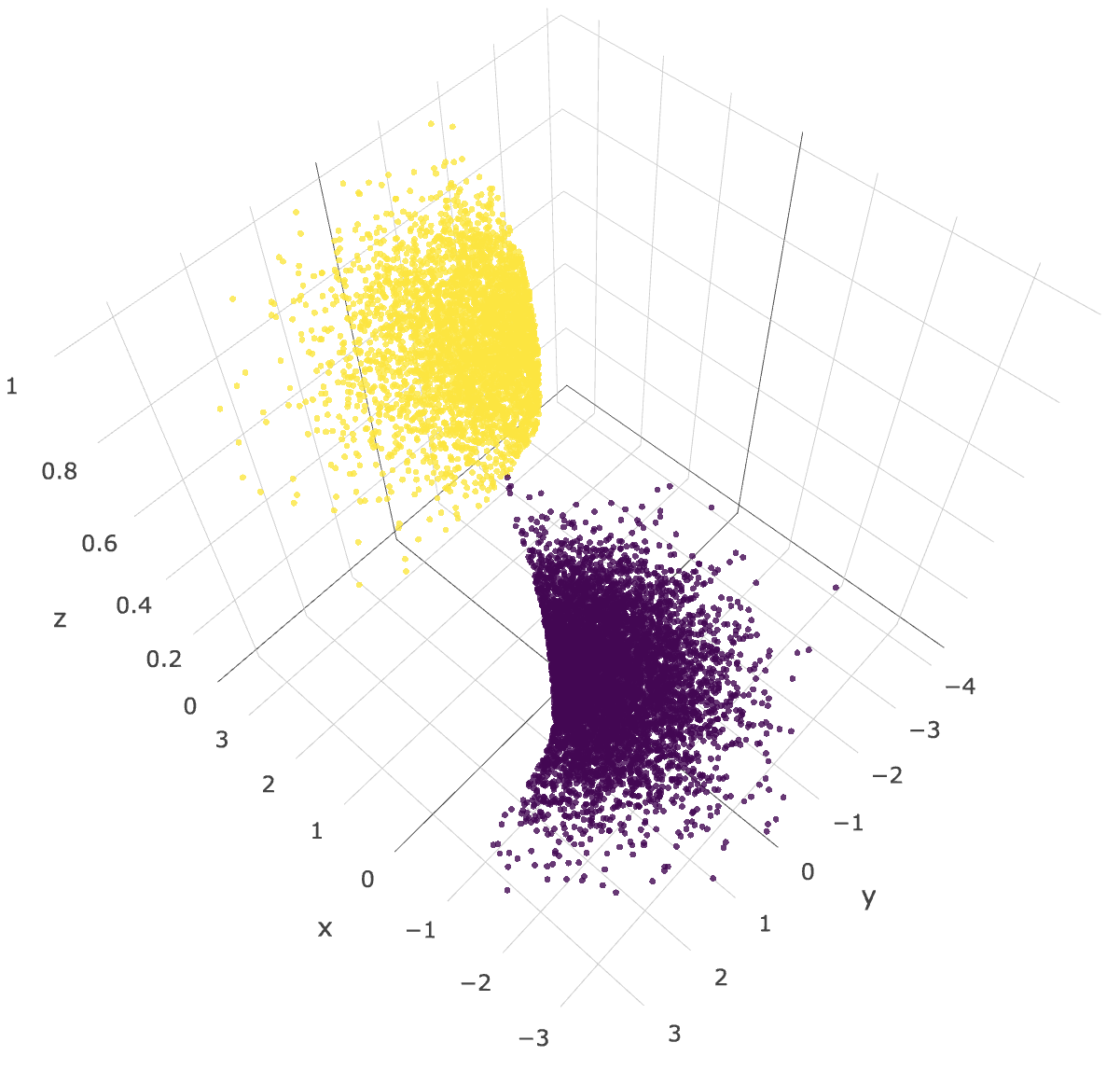}
\caption{Sample distribution} \label{fig:toy_sample}
\end{subfigure}
\hskip -0.3in
\begin{subfigure}{.6\textwidth}
  \centering
  \includegraphics[height=2.5in]{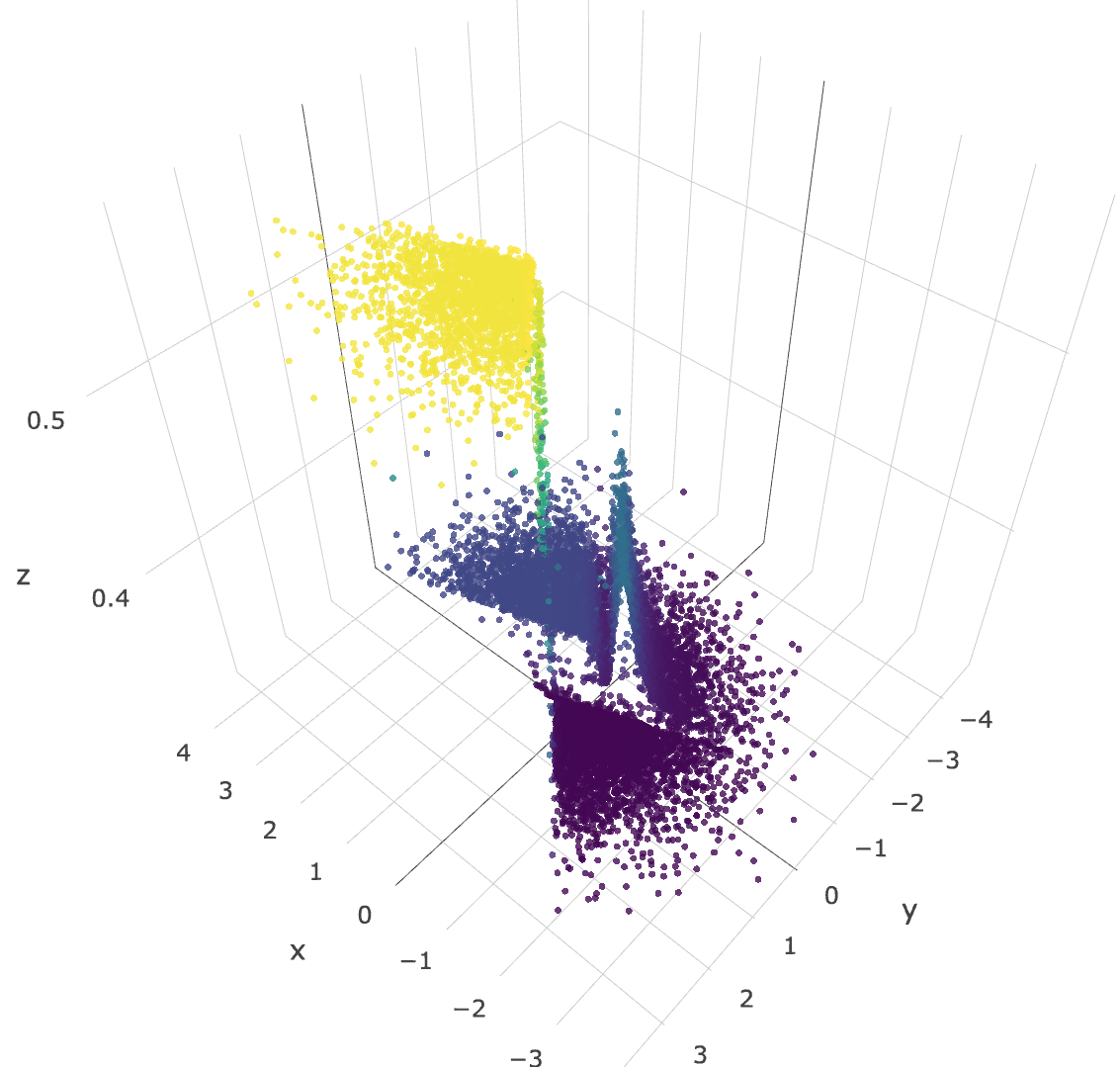}
\caption{Predicted labels by the sharp minimum}  \label{fig:toy_sharp}
\end{subfigure}
\begin{subfigure}{.6\textwidth}
  \centering
  \includegraphics[height=2.5in]{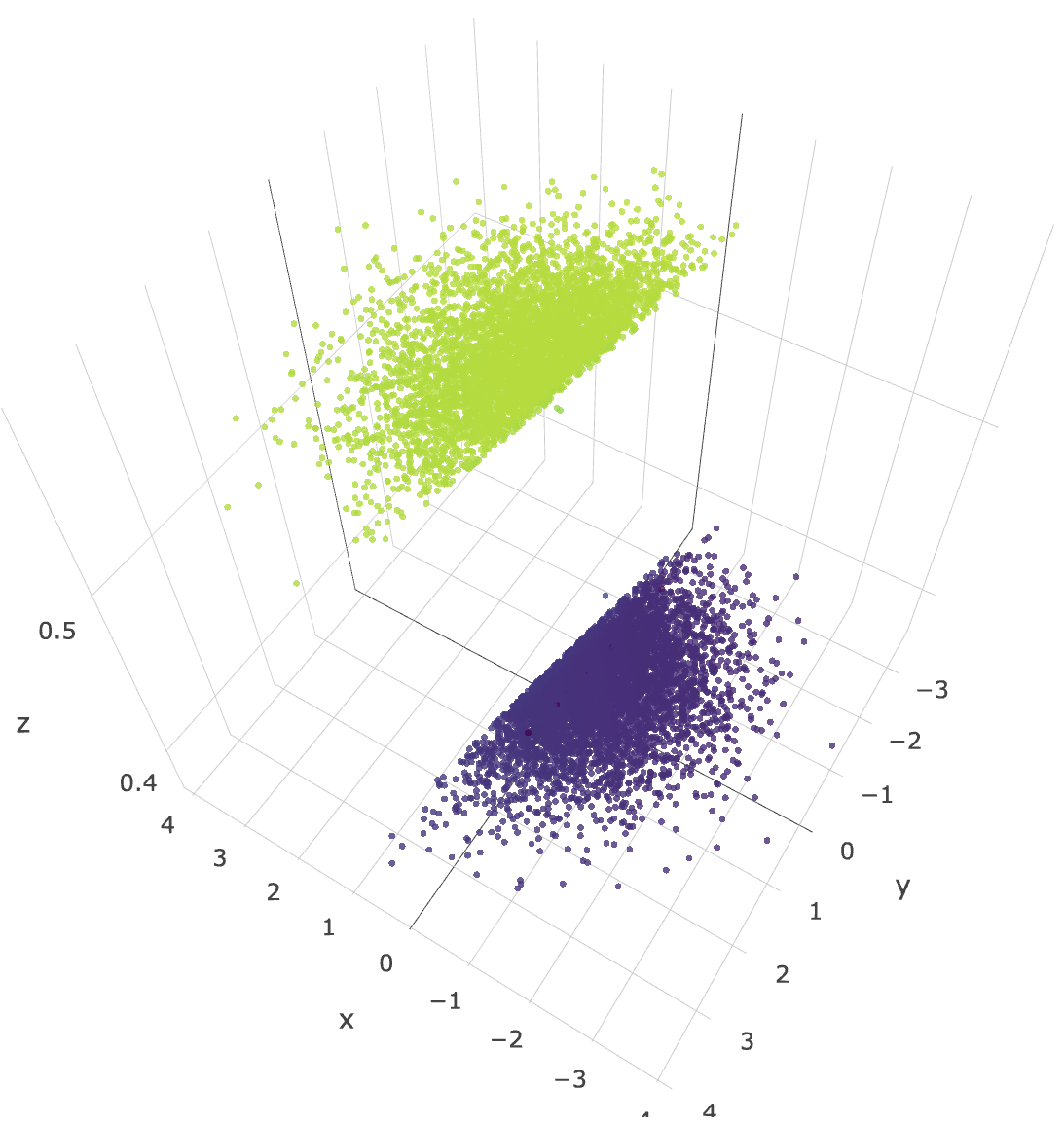}
\caption{Predicted labels by the flat minimum} \label{fig:toy_flat}
\end{subfigure}
\caption{Loss Landscape and Predicted Labels of a $5$-layer MLP with $2$ parameters.}
\end{figure}

\section{Sharp Minimum v.s. Flat Minimum - A Toy Example}

Let us start with a toy example to demonstrate different behaviors of local optima. For training, we construct a small 2-dimensional sample set from a mixture of $3$ Gaussians, and then binarize the labels by thresholding them from their median value. The sample distribution is shown in Figure \ref{fig:toy_sample}. Then we use a $5$-layer MLP model with sigmoid as the activation and cross entropy as the loss for training and prediction. The variables from different layers are shared so that the model only has two free parameters $w_1$ and $w_2$. 

The model is trained using $100$ samples. Fixing the samples, we plot the loss function with respect to the model variables $\hat{L}(w_1, w_2)$, as shown in Figure \ref{fig:toy_metrics}. Many local optima are observed even in this simple two-dimensional toy example. In particular a sharp one, marked by the vertical green line, and a flat one, marked by the vertical red line. The colors on the loss surface display the values of the generalization metric scores (pacGen), which we will define in section \ref{sec:SharpMetric}. Smaller metric value indicates better generalization power. 

As displayed in the figure, the metric score around the global optimum, indicated by the vertical green bar, is high, suggesting possible poor generalization capability as compared to the local optimum indicated by the red bar. We also plot a plane on the bottom of the figure. The color projected on the bottom plane indicates an approximated generalization bound, which considers both the loss and the generalization metric.\footnote{the bound was approximated with $\eta=39$ using inequality (\ref{eqn:uniform-eta-bound})} The local optimum indicated by the red bar, though has a slightly higher loss, has a similar overall bound compared to the ``sharp" global optimum.

On the other hand, fixing the parameter $w_1$ and $w_2$, we may also plot the labels predicted by the model given the samples. Here we plot the prediction from both the sharp minimum (Figure \ref{fig:toy_sharp}) and the flat minimum (Figure \ref{fig:toy_flat}). The sharp minimum, even though it approximates the true label better, has some complex structures in its predicted labels, while the flat minimum seems to produce a simpler classification boundary.

While it is easy to make observations on toy examples, it is less straight-forward to make a quantitative statement when the model parameters and the number of training samples grow. In the following sections we try connect the local smoothness of the solution and model generalization capability. Section \ref{sec:prelim} briefly introduces some preliminaries on the learning theory. Section \ref{sec:assump} talks about the assumptions and intuitions on how the model perturbation is related to the generalization as well as the Hessian of the solution. Section \ref{sec:pert-bound} dives into two specific types of perturbations: uniform and truncated Gaussian. Section \ref{sec:reparam} discusses the effect of re-parameterization on the proposed bound. Some empirical approximations and experiments are shown in Section \ref{sec:SharpMetric} and \ref{sec:Algo}.

\section{Model Generalization Theory}\label{sec:prelim}

We consider the general machine learning scenario. Suppose we have a labeled data set $\mathcal{S}=\{s_i=(x_i,y_i)\mid i\in\{1,\dots, n\}, x_i\in \mathbb{R}^d, y_i\in \{0, 1\}^k\}$, where $(x_i,y_i)$ are sampled i.i.d. from a distribution $x_i,y_i\sim \mathfrak{D}_s$. We try to learn a function $f\in \mathfrak{F}: \mathcal{X}\rightarrow \mathcal{Y}$, such that the expected loss 
$$L(f) = \mathbb{E}_{x,y\sim \mathfrak{D}_s}l(f,x,y)$$ 
is small, where $l: \mathfrak{F}\times \mathcal{X}\times \mathcal{Y}\rightarrow \mathbb{R}^+$ is the loss function.

Since we do not know the distribution $\mathfrak{D}_s$, the expected loss $L(f)$ is hard to calculate directly. Instead usually the empirical loss $$\hat{L}(f) = \frac{1}{n}\sum_{i=1}^n l(f, x_i, y_i)$$ is evaluated during the training procedure. 

\subsection{Rademacher Complexity}
Minimizing the empirical loss 
$$f^\ast = \arg\min_{f\in \mathfrak{F}} \hat{L}(f)$$ 
may lead to issues such as overfitting. In general, by the law of large number, for a fixed function $f\in \mathfrak{F}$, the empirical loss converges almost surely to the expected loss. However, when $f$ is not fixed, i.e., $f$ depends on the samples, and the number of samples is finite, classical learning theory suggests that the gap between the expected loss and the empirical loss is bounded by the sum of the Rademacher complexity and a concentration tail \citep{Shalev-Shwartz2014}. The Rademacher complexity is defined as 
$$\mathcal{R}_n(\mathfrak{F}, \mathfrak{D}_s) = \mathbb{E}_{x_i, y_i\sim \mathfrak{D}_s}\left[\mathbb{E}_{\epsilon} \sup_{f\in \mathfrak{F}} \frac{1}{n}\sum_{i=1}^n \epsilon_i l(f(x_i), y_i)\right],$$
where $\epsilon_i$s are i.i.d. Rademacher random variables. 

Note the Rademacher complexity is only related to the function space $\mathfrak{F}$, the sample distribution $\mathfrak{D}_s$ and the number of samples $n$. This seems to suggest when the function class is very complex, the gap between the empirical loss and the expected loss will be large.  
Though the learning theory based on Rademacher complexity can explain the overfitting effect to some extent, for example, when the hypothesis space is overly complex, the generalization tends to be worse, it is not easy to explain some well-known empirical observations in today's deep learning experiments including:
\begin{itemize}
\item Over-parameterization. 

The hypothesis space of a deep learning network can easily get rich enough to represent any function on a finite sample set \citep{Zhang2017}. According to the bound based on the Rademacher complexity, the network may tend to overfit. However empirically those deep models generalize well. 

\item Different generalization behaviors for different local optima.

The generalization bound based on Rademacher complexity holds uniformly for all hypothesis in the function class. On the other hand, it does not distinguish the generalization capabilities among different solutions. Obviously, there are ``simple" solutions even if the whole function space is complex. 
\end{itemize}

In this draft we will focus on the second empirical observations and give, to the best of our knowledge, a first explanation on behaviors of different local optima.

\subsection{PAC-Bayes}

Another line of theory discussing model generalization is PAC-Bayes \citep{Mcallester03} \citep{Mcallester1998} \citep{Mcallester1999} \citep{Langford2002}. The PAC-Bayes paradigm further assumes probability measures over the function class. In particular, it assumes a ``posterior" distribution $\mathfrak{D}_f$ as well as a ``prior" distribution $\pi_f$ over the function class $\mathfrak{F}$. In this way the function is assumed to be sampled from a ``posterior" distribution over $\mathfrak{F}$. As a consequence the expected loss is in terms of both the random draw of samples as well as the random draw of functions:
$$L(\mathfrak{D}_f, \mathfrak{D}_s) = \mathbb{E}_{f\sim \mathfrak{D}_f} \mathbb{E}_{x,y\sim \mathfrak{D}_s} l(f, x,y).$$
Correspondingly, the empirical loss in the PAC-Bayes paradigm is the expected loss over the draw of functions from the posterior:
$$\hat{L}(\mathcal{S}) = \mathbb{E}_{f\sim \mathfrak{D}_f} \frac{1}{n}\sum_{i=1}^n l(f, x_i, y_i).$$

PAC-Bayes theory suggests the gap between the expected loss and the empirical loss is bounded by a term that is related to the KL divergence between $\mathfrak{D}_f$ and $\pi_f$ \citep{Mcallester1999} \citep{Langford2002}. In particular, if the function $f$ is parameterized as $f(w)$ with $w\in \mathcal{W}$, when $\mathfrak{D}_w$ is perturbed around any $w$, we have the following PAC-Bayes bound \citep{SeldinLS2012} \citep{Seldin2011} \citep{Neyshabur2017} \citep{Neyshabur2017a}:

\begin{theorem}
[PAC-Bayes-Hoeffding Perturbation] Let $l(f, x,y)\in [0,1]$, and $\pi$ be any fixed distribution over the parameters $\mathcal{W}$. For any $\delta>0$ and $\eta>0$, with probability at least $1-\delta$ over the draw of $n$ samples, for any $w$ and any random perturbation $u$,
\begin{align}
\mathbb{E}_u [L(w+u)]\leq \mathbb{E}_u [\hat{L}(w+u)] + \frac{KL(w+u||\pi) + \log \frac{1}{\delta}}{\eta} + \frac{\eta}{2n}\label{eqn:pac-hoeffding}
\end{align}
\end{theorem}

One may further optimize $\eta$ to get a bound that scales approximately as $\mathbb{E}_u [L(w+u)]\lesssim\mathbb{E}_u [\hat{L}(w+u)] + 2\sqrt{\frac{KL(w+u||\pi) + \log \frac{1}{\delta}}{2n}}$ \citep{Seldin2011}. \footnote{Since $\eta$ cannot depend on the data, one has to build a grid and use the union bound.} A nice property of the perturbation bound (\ref{eqn:pac-hoeffding}) is it connects the generalization with the local properties around the solution $w$ through some perturbation $u$ around $w$. In particular, suppose $\hat{L}(w^\ast)$ is a local optima, when the perturbation level of $u$ is small, $\mathbb{E}_u[\hat{L}(w^\ast+u)]$ tends to be small, but $KL(w^\ast+u|\pi)$ may be large since the posterior is too ``focused" on a small neighboring area around $w^\ast$, and vice versa. As a consequence, we may need to search for an ``optimal" perturbation level for $u$ so that the bound is minimized.

\section{Local Smoothness Assumptions}\label{sec:assump}

\cite{Keskar16} investigate the local structures of the converged points for deep learning networks, and find that empirically the ``sharpness" of the minima is closely related to the generalization property of the classifier. The sharp minimizers, which led to lack of generalization ability, are characterized by a significant number of large positive eigenvalues in $\nabla^2f(x)$. In particular, they propose a local sharpness metric:
\begin{definition} [Sharpness Metric] \citep{Keskar16} Given $x\in \mathbb{R}^m$, $\epsilon > 0$ and $A\in \mathbb{R}^{m\times p}$, the $(C_\epsilon, A)$-sharpness of $f$ at $x$ is defined as:
\begin{align}
\phi_{x,f}(\epsilon, A) := \frac{\left(\max_{y\in C_\epsilon} f(x+A y)\right) - f(x)}{1+f(x)} \times 100
\end{align}
where $C_\epsilon = \left\{z\in \mathbb{R}^p:-\epsilon(|(A^+x)_i|+1)\leq z_i\leq \epsilon(|(A^+x)_i|+1), ~~\forall i\in\{1,2,\dots, p\}\right\}$, and $A^+$ is the pseudo inverse of $A$.
\end{definition}
Other variants of the model generalization metrics are also proposed by \cite{ChaudhariCSL16} and \cite{novak2018sensitivity}.

\cite{Neyshabur2017} suggests an ``expected sharpness" based on the PAC-Bayes bound:
\begin{align}
E_{u\sim N(0,\sigma^2)^m}[\hat{L}(w+u)] - \hat{L}(w)\label{eqn:sharp}
\end{align}
They also point out the sharpness itself may not be enough to determine the generalization capability, but combining scales with sharpness one may get a control of the generalization. Similar connections are also found by \cite{Dziugaite2017}.

\subsection{Smoothness Assumption over Hessian}

While some researchers have discovered empirically the generalization ability of the models is related to the second order information around the local optima, to the best of our knowledge there is no work on how to connect the Hessian matrix $\nabla^2 \hat{L}(w)$ with the model generalization. In this section we introduce the assumption about the second-order smoothness, which is later used in our generalization bound.

\begin{definition}[Hessian Lipschitz]

A twice differentiable function $f(\cdot)$ is $\rho$-Hessian Lipschitz if:
\begin{align}
\forall w_1, w_2, \|\nabla^2 f(w_1) - \nabla^2 f(w_2)\|\leq \rho \|w_1-w_2\|,\label{eqn:lipHess}
\end{align}
\end{definition}\label{def:lipHess}
where $\|\cdot\|$ is the operator norm.

The Hessian Lipschitz condition has been used in the numeric optimization community to model the second-order smoothness \citep{Nesterov2006} \citep{Zhu2014}. For the deep models it could be unrealistic to assume the Hessian Lipschitz condition holds for all $w\in \mathcal{W}$. Instead we make a local Hessian Lipschitz assumption:

\begin{definition}[Local Hessian Lipschitz]
Function $\hat{L}(w)$ is $\rho$-Hessian Lipschitz in $Neigh_{\gamma, \epsilon}(w)$, where 
$$Neigh_{\gamma, \epsilon}(w)=\{v\mid |v_i-w_i|\leq \gamma|w_i|+\epsilon~~\forall i\}$$ 
is a neighborhood around $w$ defined by two positive constants $\gamma$ and $\epsilon$.
\end{definition}

To simplify the notation in the draft we denote $\kappa_{\gamma, \epsilon}(w_i) = \gamma |w_i|+\epsilon$.

\subsection{Connecting Generalization and Hessian}

Suppose the empirical loss function $\hat{L}(w)$ satisfies the local Hessian Lipschitz condition, then by Lemma $1$ in \citep{Nesterov2006}, the perturbation of the function around a fixed point can be bounded by terms up to the third-order,
\begin{align}
\hat{L}(w+u) \leq \hat{L}(w) + \nabla \hat{L}(w)^T u + \frac{1}{2}u^T \nabla^2 \hat{L}(w) u + \frac{1}{6}\rho\|u\|^3~~~~\forall u ~~s.t.~~ w+u\in Neigh_{\gamma,\epsilon}(w)\label{eqn:lsmooth}
\end{align}

For perturbations with zero expectation, i.e., $\mathbb{E}[u]=0$, the linear term in (\ref{eqn:lsmooth}), $\mathbb{E}_u [\nabla^2\hat{L}(w)^T u]=0$. Because the perturbation $u_i$ for different parameters are independent, the second order term can also be simplified.
\begin{align}
\mathbb{E}_u \Big[\frac{1}{2}u^T \nabla^2 \hat{L}(w) u \Big] = \frac{1}{2}\sum_i \nabla^2_i \hat{L}(w) \mathbb{E}[u_i^2],\label{eqn:2ndexp}
\end{align}
where $\nabla^2_i$ is simply the $i$-th diagonal element in Hessian. The following lemma is straight-forward given (\ref{eqn:pac-hoeffding}),(\ref{eqn:lsmooth}), and (\ref{eqn:2ndexp}).

\begin{lemma}
Suppose the loss function $l(f, x,y)\in [0,1]$. Let $\pi$ be any distribution on the parameters that is independent from the data.  For any $\delta>0$ and $\eta>0$, with probability at least $1-\delta$ over the draw of $n$ samples, for any $w$ such that $\hat{L}(w)$ satisfies the local $\rho$-Hessian Lipschitz condition in $Neigh_{\gamma, \epsilon}(w)$, and any random perturbation $u$, s.t., $|u_i|\leq \kappa_{\gamma,\epsilon}(w_i)~~\forall i$,  $\mathbb{E}[u] = 0$, $u_i$ and $u_j$ are independent for any $i\neq j$, we have
\begin{align}
\mathbb{E}_u [L(w+u)]\leq \hat{L}(w) +\frac{1}{2} \sum_i \nabla_i^2 \hat{L}(w)\mathbb{E} [u_i^2] + \frac{\rho}{6}\mathbb{E}[\|u\|^3] + \frac{KL(w+u||\pi) + \log \frac{1}{\delta}}{\eta} + \frac{\eta}{2n}\label{eqn:lemma1}
\end{align}
where $\nabla_i$ is the $i$-th diagonal element of $\nabla \hat{L}(w)$.
\end{lemma}\label{lemma:pac2nd}

Note by extrema of the Rayleigh quotient, the quadratic term on the right hand side of inequality (\ref{eqn:lsmooth}) is further bounded by
\begin{align}
u^T \nabla^2 \hat{L}(w) u\leq \lambda_{max}(\nabla^2 \hat{L}(w)) \|u\|^2. \label{eqn:rayleigh}
\end{align}
This is consistent with the empirical observations of \cite{Keskar16} that the generalization ability of the model is related to the eigenvalues of $\nabla^2 \hat{L}(w)$. The inequality (\ref{eqn:rayleigh}) still holds even if the perturbations $u_i$ and $u_j$ are correlated. We add another lemma about correlated perturbations in Appendix (Lemma \ref{lemma:app-1}). 

\subsection{Tradeoff between Sharpness Metric and Generalization Power}
If we look at the right hand side of the inequality (\ref{eqn:lemma1}), and compare it with (\ref{eqn:sharp}) \citep{Neyshabur2017}, we see
\begin{align}
\mathbb{E}_u \hat{L}(w+u) - \hat{L}(w) \leq  \mathcal{M}(w, \mathcal{D}_u) = \frac{1}{2} \sum_i \nabla^2_i \hat{L}(w)\mathbb{E} [u_i^2] + \frac{\rho}{6}\mathbb{E}[\|u\|^3]
\end{align}

\begin{figure}
  \centering
	\resizebox{0.9\textwidth}{0.3\textwidth}{\includegraphics{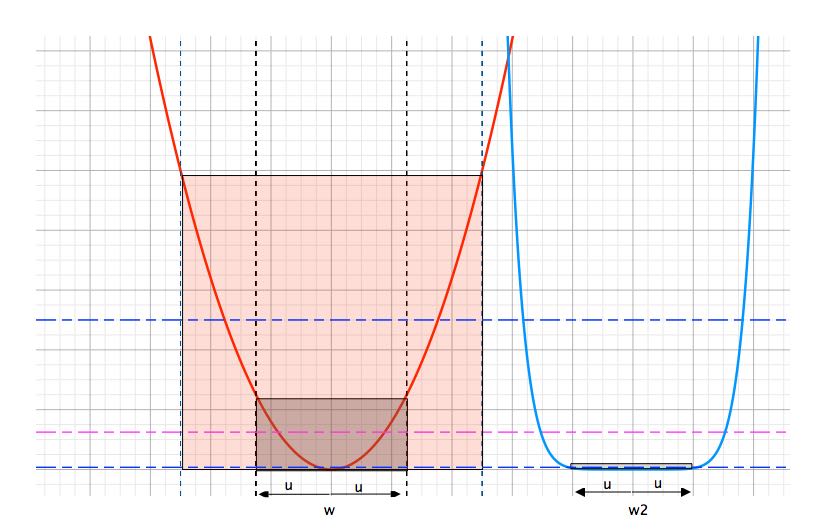}}
  \caption{Sharpness Metric for $\hat{L}(w)$, $1$-dimensional case. Fixing the perturbation level, larger $\nabla^2 \hat{L}(w)$ leads to larger $\mathcal{M}(w, \mathcal{D}_u)$.}
\label{fig:sharpvsflat}
\end{figure}

$\mathcal{M}(w, \mathcal{D}_u)$ can be interpreted as the sharpness metric of the empirical loss. It is closely related to the Hessian $\nabla^2 L(w)$, but it is also related to the perturbation distributions. Figure (\ref{fig:sharpvsflat}) shows when the perturbation is fixed how $\nabla^2\hat{L}(w)$ can affect the term $\mathbb{E}_u\hat{L}(w)$.

The other term 
\begin{align}
\mathcal{G}_{\delta, n}(\eta, \mathcal{D}_{w+u}, \pi) = \frac{KL(w+u||\pi) + \log \frac{1}{\delta}}{\eta} + \frac{\eta}{2n}
\end{align}
is related to the model generalization power in the original PAC-Bayes bound. 

Ideally we would like both $\mathcal{M}(w, \mathcal{D}_u)$ and $\mathcal{G}_{\delta, n}(\eta, \mathcal{D}_{w+u}, \pi)$ to be small for better generalization capability. However, generally the perturbation distribution that leads to small $\mathcal{M}(w, \mathcal{D}_u)$ tends to have large $\mathcal{G}_{\delta, n}(\eta, \mathcal{D}_{w+u}, \pi)$ for a given prior. As we will see in the following sections, in the end we have to make trade-offs between the two terms. \\

\section{Bounded Perturbations}\label{sec:pert-bound}

Adding noise to the model for better generalization has proven successful both empirically and theoretically \citep{Zhu2018} \citep{Hoffer2017} \citep{Jastrzebski2017} \citep{Dziugaite2017} \citep{Novak2018}. Instead of only minimizing the empirical loss, \citep{Langford2001} and \citep{Dziugaite2017} assume different perturbation levels on different parameters, and minimize the generalization bound led by PAC-Bayes for better model generalization. However how to connect the noise distribution with the local optima structures, for example, $\nabla^2 L(w^\ast)$, and how that is related to the generalization power have not been examined. 

Since the assumptions in Lemma (\ref{lemma:pac2nd}) are local, the distributions of interest for the perturbation are necessarily bounded. In this section we investigate two special forms of perturbations, the uniform perturbation and truncated Gaussian, and provide closed-form scale estimation for the perturbation levels. 

\subsection{Uniform Distribution}
Suppose $u_i\sim U(-\sigma_i, \sigma_i)$, and $\sigma_i\leq \kappa_{\gamma, \epsilon}(w_i)~~\forall i$. That is, the ``posterior" distribution of the model parameters are uniform distribution, and the distribution supports vary for different parameters. We also assume the perturbed parameters are bounded, i.e., $|w_i| + \kappa_{\gamma, \epsilon}(w_i)\leq \tau_i ~~\forall i$.\footnote{One may also assume the same $\tau$ for all parameters for a simpler argument. The proof procedure goes through in a similar way.} If we choose the priors $\pi$ to be $u_i\sim U(-\tau_i, \tau_i)$, and then
\begin{align}
KL(w+u||\pi) = \sum_i\log (\tau_i/\sigma_i)
\end{align}

Note $E[u_i^2] = \sigma_i^2/3$. Also we simplify the third order term in (\ref{eqn:lemma1}) by 
$$\frac{\rho}{6}\mathbb{E}[\|u\|^3]\leq\frac{\rho m^{1/2}}{6}\mathbb{E}[\|u\|_3^3]\leq \frac{\rho m^{1/2}}{6}\sum_i \kappa_{\gamma, \epsilon}(w_i)\mathbb{E}[u_i^2]=\frac{\rho m^{1/2}}{18}\sum_i \kappa_{\gamma, \epsilon}(w_i)\sigma_i^2,$$
where we use the inequality $\|u\|_2\leq m^{\frac{1}{6}}\|u\|_3$ and $m$ is the number of parameters. By Lemma (\ref{lemma:pac2nd}), we get
\begin{align}
\mathbb{E}_u [L(w+u)]\leq \hat{L}(w) +\frac{1}{6} \sum_i \nabla_i^2 L(w)\sigma_i^2 + \frac{\rho m^{1/2}}{18}\sum_i \kappa_{\gamma, \epsilon}(w_i)\sigma_i^2 + \frac{\sum_i\log \frac{\tau_i}{\sigma_i} + \log \frac{1}{\delta}}{\eta} + \frac{\eta}{2n}\label{eqn:uniform-bound}
\end{align}

If we assume $\hat{L}(w)$ is locally convex  around $w^\ast$ so that $\nabla^2_i\hat{L}(w^\ast)\geq 0$ for all $i$. Solve for $\sigma$ that minimizes the right hand side, and we have the following lemma:
\begin{lemma}\label{lemma:uniform-pertb-bound}
Suppose the loss function $l(f, x,y)\in [0,1]$, and model weights are bounded $|w_i| + \kappa_{\gamma, \epsilon}(w_i)\leq \tau_i ~~\forall i$. For any $\delta>0$ and $\eta$, with probability at least $1-\delta$ over the draw of $n$ samples, for any $w^\ast\in \mathbb{R}^m$ such that $\hat{L}(w)$ is locally convex in $Neigh_{\gamma, \epsilon}(w^\ast)$ and $\hat{L}(w)$ satisfies the local $\rho$-Hessian Lipschitz condition in $Neigh_{\gamma, \epsilon}(w^\ast)$,
\begin{align}
\mathbb{E}_u [L(w^\ast+u)]\leq \hat{L}(w^\ast)  + \frac{m/2 + \sum_i \log \frac{\tau_i}{\sigma_i^\ast} + \log\frac{1}{\delta}}{\eta} + \frac{\eta}{2n} \label{eqn:uniform-eta-bound}
\end{align}
where $u_i\sim U(-\sigma_i^\ast, \sigma_i^\ast)$ are i.i.d. uniformly perturbed random variables, and
\begin{align}
\sigma_i^\ast(w^\ast,\eta, \gamma) =\min\left(\sqrt{\frac{1}{\eta(\nabla^2_i L(w^\ast)/3 + \rho m^{1/2}\kappa_{\gamma, \epsilon}(w_i^\ast)/9))}}, \kappa_{\gamma, \epsilon}(w_i^\ast)\right).\label{eqn:uniform-sigma}
\end{align}
\end{lemma}

In our experiment, we simply treat $\eta$ as a hyper-parameter. Other other hand, one may further build a weighted grid over $\eta$ and optimize for the best $\eta$ \citep{Seldin2011}. In this way we reach the following theorem:
\begin{theorem}\label{theorem:uniform-pertb-bound-refine}
Under the conditions of Lemma \ref{lemma:uniform-pertb-bound}, for any $\delta>0$, with probability at least $1-\delta$ over the draw of $n$ samples, for any $w^\ast\in \mathbb{R}^m$ such that in $Neigh_{\gamma, \epsilon}(w^\ast)$, $\hat{L}(w)$ is locally convex and satisfies the local $\rho$-Hessian Lipschitz condition, 
\begin{align}
\mathbb{E}_u [L(w^\ast+u)]
&\leq \hat{L}(w^\ast) + O\left(\sqrt{\frac{m+\sum_i \log \frac{\tau_i}{\sigma_i^\ast} + \log \frac{1}{\delta}}{n}}\right)\nonumber
\end{align}
where $u_i\sim U(-\sigma_i^\ast, \sigma_i^\ast)$ are i.i.d. uniformly perturbed random variables, and
\begin{align}
\sigma_i^\ast(w^\ast,\eta, \gamma) =\min\left(\sqrt{\frac{1}{\sqrt{m n}(\nabla^2_i \hat{L}(w^\ast)/3 + \rho m^{1/2}\kappa_{\gamma, \epsilon}(w_i^\ast)/9)}}, \kappa_{\gamma, \epsilon}(w_i^\ast)\right)\label{eqn:app-uniform-sigma-2}
\end{align}
\end{theorem}

Please see the appendix for the details of the proof.

\subsection{Truncated Gaussian}
Because the Gaussian distribution is not bounded but Lemma (\ref{lemma:pac2nd}) requires bounded perturbation, we first truncate the distribution. The procedure of truncation is similar to the proof in \citep{Neyshabur2017a} and \citep{Mcallester03}. 

Let $u\sim N(0, \Sigma)$, where $\Sigma$ is a diagonal covariance matrix. Denote the truncated Gaussian as $N_{\gamma,\epsilon}(0,\Sigma)$. If $\tilde{u}\sim N_{\gamma,\epsilon}(0,\Sigma)$ then
\begin{align}
\mathbb{P}_{\gamma, \epsilon}(\tilde{u}) = \frac{1}{Z}
\left\{
\begin{array}{c l}	
     p(u) & \mathrm{if}~~|u_i| < \kappa_{\gamma, \epsilon}(w_i)~~\forall i\\
     0 & o.w.
\end{array}\right.\label{eqn:gauss-trunc}
\end{align}

If $~~\forall i ~~\sigma_i < \frac{\kappa_{\gamma, \epsilon}(w_i)}{\sqrt{2}\mathrm{erf}^{-1}(\frac{1}{2m})}$, by union bound $Z\geq 1/2$. Here $\mathrm{erf}^{-1}$ is the inverse Gaussian error function defined as $\mathrm{erf}(x)=\frac{2}{\sqrt{\pi}}\int_0^xe^{-t^2}dt$, and $m$ is the number of parameters. Following a similar procedure as in the proof of Lemma 1 in \citep{Neyshabur2017a},
\begin{align}
KL(w+\tilde{u}||\pi)\leq 2(KL(w+u||\pi)+1)
\end{align}
Suppose the coefficients are bounded such that $\sum_i w_i^2\leq \tau$, where $\tau$ is a constant. Choose the prior $\pi$ as $N(0, \tau I)$, and we have
\begin{align}
KL(w+u||\pi) \leq \frac{1}{2}(m\log \tau -\sum_i \log \sigma_i^2 - m + \frac{1}{\tau}\sum_i \sigma_i^2 + 1) 
\end{align}

Notice that after the truncation the variance only becomes smaller, so the bound of (\ref{eqn:lemma1}) for the truncated Gaussian becomes
\begin{align}
\mathbb{E}_u [L(w+\tilde{u})]\leq \hat{L}(w) + &\frac{1}{2} \sum_i \nabla_i^2 L(w)\sigma_i^2 + \frac{\rho m^{1/2}}{6} \sum_i \kappa_{\gamma, \epsilon}(w_i)\sigma_i^2 \nonumber\\&+\frac{m\log \tau -\sum_i \log \sigma_i^2 - m + \frac{1}{\tau}\sum_i \sigma_i^2 + 1+ 2\log\frac{1}{\delta} }{2\eta} + \frac{\eta}{2n}\label{eqn:gauss-bound}
\end{align}

Again when $\hat{L}(w)$ is convex around $w^\ast$ such that $\nabla^2 \hat{L}(w^\ast)\geq 0$, solve for the best $\sigma_i$ and we get the following lemma:

\begin{lemma}\label{lemma:gauss-pertb-bound}
Suppose the loss function $l(f, x,y)\in [0,1]$, and model weights are bounded $\sum_i w_i^2\leq \tau$. For any $\delta>0$ and $\eta$, with probability at least $1-\delta$ over the draw of $n$ samples, for any $w^\ast\in \mathbb{R}^m$ such that in $Neigh_{\gamma, \epsilon}(w^\ast)$, $\hat{L}(w)$ is convex and satisfies the local $\rho$-Hessian Lipschitz condition,
%as defined in (\ref{eqn:gauss-trunc}), 
\begin{align}
\mathbb{E}_u [L(w^\ast+\tilde{u})]\leq \hat{L}(w^\ast)  + \frac{m\log \tau -\sum_i \log \sigma_i^2 + 1+ 2\log\frac{1}{\delta} }{2\eta} + \frac{\eta}{2n} 
\end{align}
where $\tilde{u}\sim N_{\gamma,\epsilon}(0,\Sigma^\ast)$ are random variables distributed as truncated Gaussian,   
\begin{align}
\sigma_i^\ast = \min\left(\sqrt{\frac{1}{\eta\nabla_i^2 \hat{L}(w^\ast) + \frac{\rho\eta m^{1/2}}{3}\kappa_{\gamma, \epsilon}(w_i^\ast) + \frac{1}{\tau}}}, \frac{\kappa_{\gamma, \epsilon}(w_i^\ast)}{\sqrt{2}\mathrm{erf}^{-1}(\frac{1}{2m})}\right)
\end{align} and $\sigma_i^{\ast 2}$ is the $i$-th diagonal element in $\Sigma^\ast$.
\end{lemma}

Again We have an extra term $\eta$, which may be further optimized over a grid to get a tighter bound. In our algorithm we treat $\eta$ as a hyper-parameter instead.

\section{On the Re-parameterization of RELU-MLP}\label{sec:reparam}

\cite{Dinh2017} points out the spectrum of $\nabla^2 \hat{L}$ itself is not enough to determine the generalization power. One particular example is the multiple layer perceptron with RELU as the activations (RELU-MLP). For a two-layer RELU-MLP, denote $w^1$, and $w^2$ as the linear coefficients for the first and second layer. Clearly 
\begin{align}
\hat{L}(w^1, w^2) = \hat{L}(\alpha w^1, \alpha^{-1}w^2)
\end{align}
If cross entropy (negative log likelihood) is used as the loss function, under certain regularization conditions, if $p(x,y) = f(x,w^\ast)[y]$, i.e., $w^\ast$ is the ``true" parameter of the sample distribution, the change in Hessian to re-parameterization can be calculated as the outer product of the gradients, in this case
\begin{align}
\nabla^2\hat{L}(\alpha w^1, \alpha^{-1}w^2) = \begin{bmatrix}\alpha^{-1} I_{m^1} & 0\\0 &\alpha I_{m^2}\end{bmatrix}\nabla^2\hat{L}(w^1, w^2) \begin{bmatrix}\alpha^{-1} I_{m^1} & 0\\0 &\alpha I_{m^2}\end{bmatrix}
\end{align}

In general our bound does not assume the loss function to be cross entropy loss. Also we do not assume the model is RELU-MLP. As a result we would not expect our bound stays exactly the same during the re-parameterization. 

On the other hand, the optimal perturbation levels in our bound scales inversely during the scaling of parameters, so the bound only changes approximately with a speed of logarithmic factor. According to Lemma (\ref{lemma:uniform-pertb-bound}) and (\ref{lemma:gauss-pertb-bound}), if we use the optimal $\sigma^\ast$ on the right hand side of the bound, $\nabla^2 \hat{L}(w)$, $\rho$, and $w^\ast$ are all behind the logarithmic terms. As a consequence, for RELU-MLP, if we do the re-parameterization trick as in \cite{Dinh2017}, the change of the bound is small.\\\\
\textbf{Disclaim: Section \ref{sec:SharpMetric} and \ref{sec:Algo} will be heuristic-based experiments and approximations. They are not rigorous.}

\section{An Approximate Generalization Metric}\label{sec:SharpMetric}%\label{sec:metric-paGEN}

Assuming $\hat{L}(w)$ is locally convex around $w^\ast$, so that $\nabla^2_i \hat{L}(w^\ast)\geq 0$ for all $i$. If we look at Lemma \ref{lemma:uniform-pertb-bound}, for fixed $m$ and $n$, the only relevant term is $\sum_i \log \frac{\tau_i}{\sigma_i^\ast}$. Replacing the optimal $\sigma^\ast$, and using $\tau_i\sim |w_i| + \kappa_{\gamma, \epsilon}(w_i)$ to approximate $\tau_i$, we come up with \textbf{PAC}-Bayes based \textbf{Gen}eralization metric, called pacGen,\footnote{Even though we assume the local convexity in our metric, in application we may calculate the metric on every points. When $\nabla^2_i \hat{L}(w^\ast) + \rho(w^\ast)\sqrt{m}\kappa_{\gamma,\epsilon}(w_i^\ast) < 0$ we simply treat it as $0$.}
\begin{align}
\Psi_{\gamma, \epsilon}(\hat{L}, w^\ast) = \sum_i \log \left(\left(|w_i^\ast| + \kappa_{\gamma, \epsilon}(w_i^\ast)\right)\max\left(\sqrt{\nabla^2_i \hat{L}(w^\ast) + \rho(w^\ast)\sqrt{m}\kappa_{\gamma, \epsilon}(w_i^\ast)}, \frac{1}{\kappa_{\gamma, \epsilon}(w_i^\ast)}\right)\right).
\end{align}

\begin{figure}
\begin{subfigure}{.5\textwidth}
  \centering
  \includegraphics[width=1.0\linewidth]{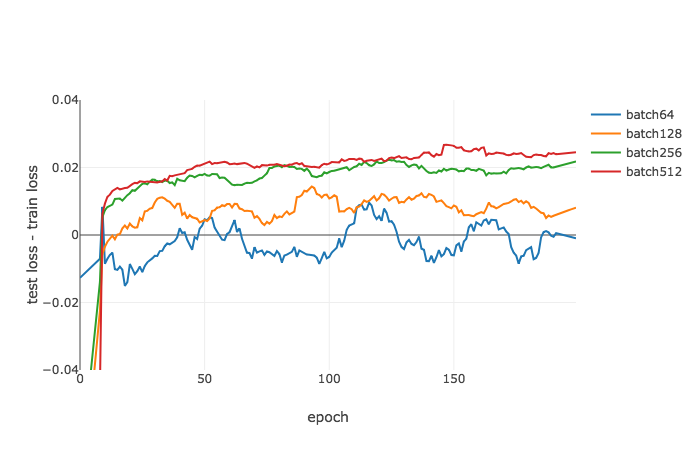}
  \caption{Test Loss - Train Loss (MNIST)}
\end{subfigure}%
\begin{subfigure}{.5\textwidth}
  \centering
  \includegraphics[width=1.0\linewidth]{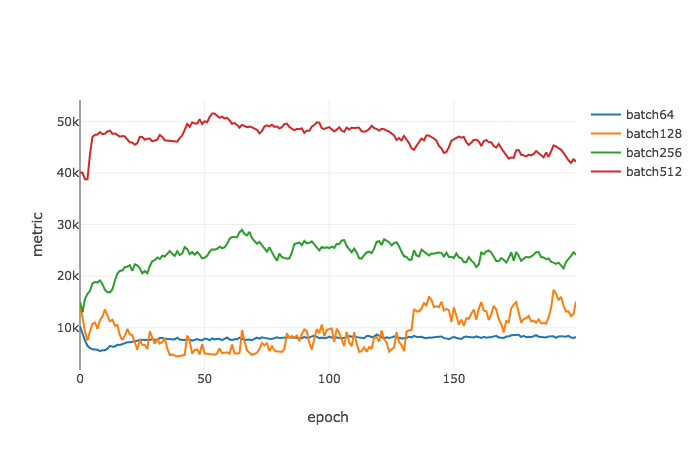}
  \caption{$\Psi_{\gamma=0.1, \epsilon=0.1}$ (MNIST)}
\end{subfigure}
\caption{Generalization gap and $\Psi_{\gamma=0.1, \epsilon=0.1}$ as a function of epochs on MNIST for different batch sizes. SGD is used as the optimizer, and the learning rate is set as $0.1$ for all configurations. As the batch size grows, $\Psi_{\gamma, \epsilon}(\hat{L}, w^\ast)$ gets larger. The trend is consistent with the true gap of losses.}
\label{fig:metric_batch_mnist}
\end{figure}

\begin{figure}
\begin{subfigure}{.5\textwidth}
  \centering
  \includegraphics[width=1.0\linewidth]{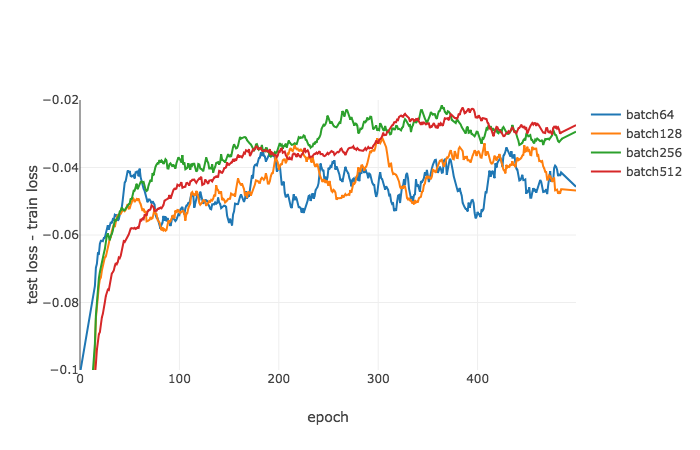}
  \caption{Test Loss - Train Loss (CIFAR-10)}
\end{subfigure}%
\begin{subfigure}{.5\textwidth}
  \centering
  \includegraphics[width=1.0\linewidth]{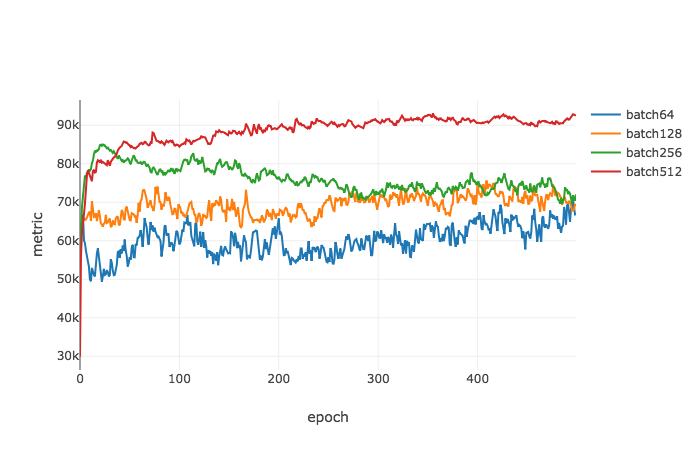}
  \caption{$\Psi_{\gamma=0.1, \epsilon=0.1}$ (CIFAR-10)}
\end{subfigure}
\caption{Generalization gap and $\Psi_{\gamma=0.1, \epsilon=0.1}$ as a function of epochs on CIFAR-10 for different batch sizes. SGD is used as the optimizer, and the learning rate is set as $0.01$ for all configurations.}
\label{fig:metric_batch_cifar}
\end{figure}

To calculate the metric on real-world data we need to estimate the diagonal elements of the Hessian $\nabla^2 \hat{L}$ as well as the Lipschitz constant $\rho$ of the Hessian. For efficiency concern we follow Adam \citep{KingmaB14} and approximate $\nabla^2_i\hat{L}$ by $(\nabla \hat{L}[i])^2$. Also we use the exponential smoothing technique with $\beta=0.999$ as in \citep{KingmaB14}. 

To estimate $\rho$, we first estimate the Hessian of a randomly perturbed model $\nabla^2\hat{L}(w+u)$\footnote{In the experiment the gradients are taken w.r.t. $w$ instead of $w+u$, and we ignore the difference between $\nabla^2_{w}\hat{L}(w+u)$ and $\nabla^2_{w+u}\hat{L}(w+u)$.}, and then approximate $\rho$ by $\rho = \max_i\frac{|\nabla_i^2 L(w + u_i) - \nabla_i^2 L(w)|}{|u_i|}$. 

We used the same model without dropout from the PyTorch example \footnote{\url{https://github.com/pytorch/examples/tree/master/mnist}}. We fix the learning rate as $0.1$ and vary the batch size for training. The gap between the test loss and the training loss, and the metric $\Psi_{\gamma, \epsilon}(\hat{L}, w^\ast)$ are plotted in Figure \ref{fig:metric_batch_mnist}. We had the same observation as in \citep{Keskar16} that as the batch size grows, the gap between the test loss and the training loss tends to get larger. Our proposed metric $\Psi_{\gamma, \epsilon}(\hat{L}, w^\ast)$ also shows the exact same trend. Note we do not use LR annealing heuristics as in \citep{Goyal2017} which enables large batch training.

\begin{figure}
\begin{subfigure}{.5\textwidth}
  \centering
  \includegraphics[width=1.0\linewidth]{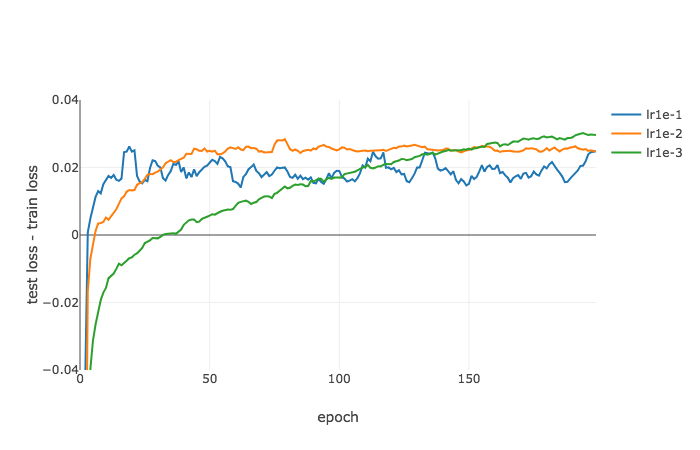}
  \caption{Test Loss - Train Loss (MNIST)}
\end{subfigure}%
\begin{subfigure}{.5\textwidth}
  \centering
  \includegraphics[width=1.0\linewidth]{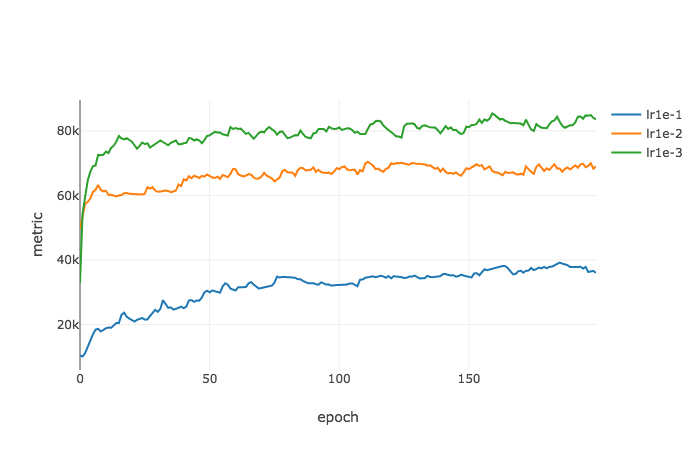}
  \caption{$\Psi_{\gamma=0.1, \epsilon=0.1}$ (MNIST)}
\end{subfigure}
\caption{Generalization gap and $\Psi_{\gamma=0.1, \epsilon=0.1}$ as a function of epochs on MNIST for different learning rates. SGD is used as the optimizer, and the batch size is set as $256$ for all configurations. As the learning rate shrinks, $\Psi_{\gamma, \epsilon}(\hat{L}, w^\ast)$ gets larger. The trend is consistent with the true gap of losses.}
\label{fig:metric_lr_mnist}
\end{figure}

Similarly we also carry out experiment by fixing the training batch size as $256$, and varying the learning rate. Figure \ref{fig:metric_lr_mnist} shows generalization gap and $\Psi_{\gamma, \epsilon}(\hat{L}, w^\ast)$ as a function of epochs. It is observed that as the learning rate decreases, the gap between the test loss and the training loss increases. And the proposed metric $\Psi_{\gamma, \epsilon}(\hat{L}, w^\ast)$ shows similar trend compared to the actual generalization gap.

\begin{figure}
\begin{subfigure}{.5\textwidth}
  \centering
  \includegraphics[width=1.0\linewidth]{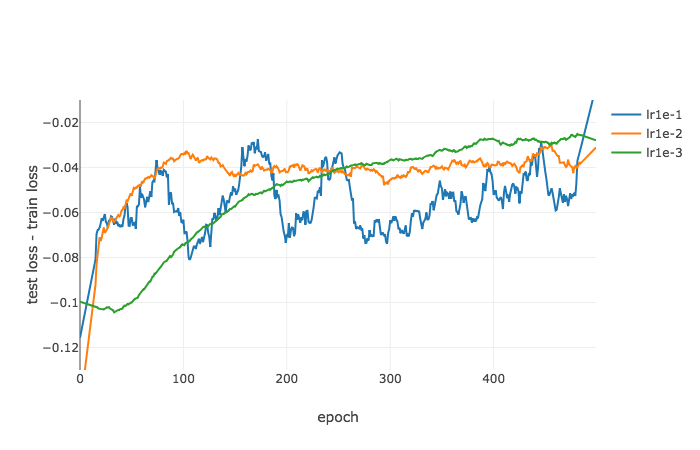}
  \caption{Test Loss - Train Loss (CIFAR-10)}
\end{subfigure}%
\begin{subfigure}{.5\textwidth}
  \centering
  \includegraphics[width=1.0\linewidth]{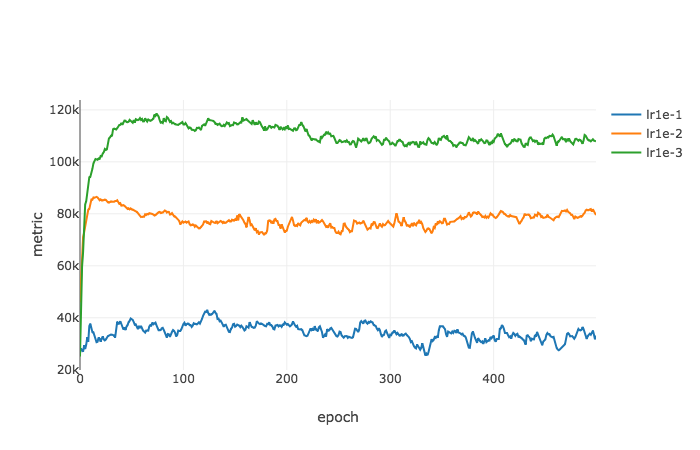}
  \caption{$\Psi_{\gamma=0.1, \epsilon=0.1}$ (CIFAR-10)}
\end{subfigure}
\caption{Generalization gap and $\Psi_{\gamma=0.1, \epsilon=0.1}$ as a function of epochs on CIFAR-10 for different learning rates. SGD is used as the optimizer, and the batch size is set as $256$ for all configurations.}
\label{fig:metric_lr_cifar}
\end{figure}

We also run the same model and experiment on CIFAR-10 \citep{CIFAR10} just to demonstrate the effectiveness of the metric. We observed similar trends on CIFAR-10 as shown in Figure \ref{fig:metric_batch_cifar} and Figure \ref{fig:metric_lr_cifar}.

\section{A Perturbed Optimization Algorithm}\label{sec:Algo}

The right hand side of (\ref{eqn:pac-hoeffding}) has $\mathbb{E}_u [\hat{L}(w+u)]$. This suggests rather than minimizing the empirical loss $\hat{L}(w)$, we should optimize the perturbed empirical loss $\mathbb{E}_u [\hat{L}(w+u)]$ instead for a better model generalization power. Adding perturbation to the model is not a new trick. Most of the perturbation-based methods \citep{Zhu2018} \citep{Hoffer2017} \citep{Jastrzebski2017} \citep{Novak2018} \citep{KhanNTLGS18} are based on heuristic techniques and improvement in applications have already been observed empirically. \cite{Dziugaite2017} first proposes to optimize for a better perturbation level from the PAC-Bayes bound, but their bound is not making use of the second order information. Also the best perturbation in \citep{Dziugaite2017} is not close-form.

In this section we introduce a systematic way to perturb the model weights based on the PAC-Bayes bound. Again we use the same exponential smoothing technique as in Adam \citep{KingmaB14} to estimate the Hessian $\nabla^2\hat{L}$. To make the algorithm efficient, we ignore the third order part in the bound (\ref{eqn:lemma1}) so that we do not have to estimate the Lipschitz constant $\rho$ of Hessian. The details of the algorithm is presented in (Algorithm \ref{Algo:1}), where we treat $\eta$ as a hyper-parameter to be optimized using the validation set.

\begin{algorithm}
\caption{Perturbed OPT}
\begin{algorithmic}[1]
\Require $\eta$, $\gamma=0.1$, $\beta_1=0.999$, $\beta_2=0.1$, $\epsilon$=1e-5.
  \State Initialization: $\sigma_i\leftarrow 0$ for all $i$. $t\leftarrow 0$, $h_0\leftarrow 0$
\For {epoch in $1,\dots, N$}
\For {minibatch in one epoch}
          \For {all $i$}
            \If {$t>0$}
            \State $\rho[i]\leftarrow \frac{|h_{t+1}[i]-h_t[i]|}{\|w_{t+1}-w_t\|}$
            \State $\kappa[i]\leftarrow \frac{\gamma}{\log(1+epoch)}|w_t[i]|+\epsilon$
            \State $\sigma_i \leftarrow \min\left(\frac{1}{\log (1+epoch)\sqrt{\eta (h_t[i] + \rho[i]\cdot \kappa[i])}  }, \kappa[i]\right) \cdot \mathbf{1}_{|g_t[i]|<\beta_2}$
           \EndIf            
          \State $u_t[i]\sim U(-\sigma_i,\sigma_i)$(sample perturbation)        
          \EndFor
  		\State $g_{t+1} \leftarrow \nabla_w \hat{L}_t(w_t + u_t)$ (get stochastic gradients w.r.t. perturbed loss)
%        \State $v_{t+1}\leftarrow\beta_1 v_t + (1-\beta_1)g_{t+1}$
        \State $h_{t+1} \leftarrow \beta_1 h_t + (1-\beta_1)g_{t+1}^2$ (update second moment estimate)
        \State $w_{t+1}\leftarrow \mathrm{OPT}(w_t)$ (update $w$ using off-the-shell algorithms)
        
     \State $t\leftarrow t+1$
\EndFor        
\EndFor
\end{algorithmic}\label{Algo:1}
\end{algorithm}

Even though in theoretical analysis $E_u[\nabla\hat{L} \cdot u]=0$, in applications, $\nabla\hat{L} \cdot u$ won't be zero especially when we only implement $1$ trial of perturbation. On the other hand, if the gradient $\nabla \hat{L}$ is close to zero, then the first order term can be ignored. As a consequence, in (Algorithm \ref{Algo:1}) we only perturb the parameters that have small gradients whose absolute value is below $\beta_2$. For efficiency issues we used a per-parameter $\rho_i$ capturing the variation of the diagonal element of Hessian. Also we decrease the perturbation level with a log factor as the epoch increases. 

\begin{figure}
\hskip -0.3in
\begin{subfigure}{.3\textwidth}
  \centering
  \includegraphics[width=55mm,height=40mm]{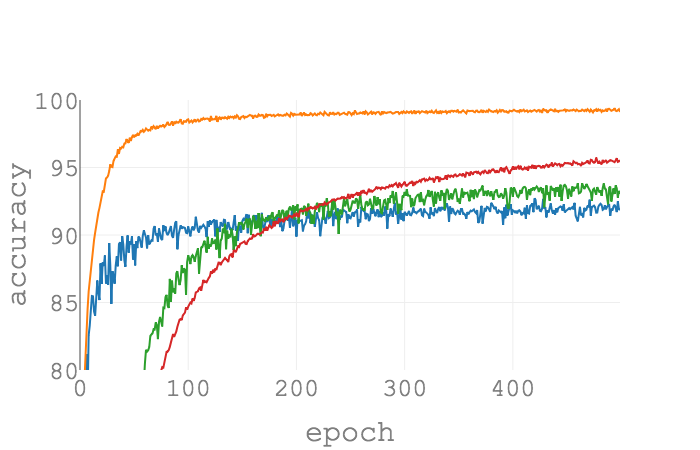}
  \caption{CIFAR-10}
\end{subfigure}%
\begin{subfigure}{.3\textwidth}
  \centering
  \includegraphics[width=55mm,height=40mm]{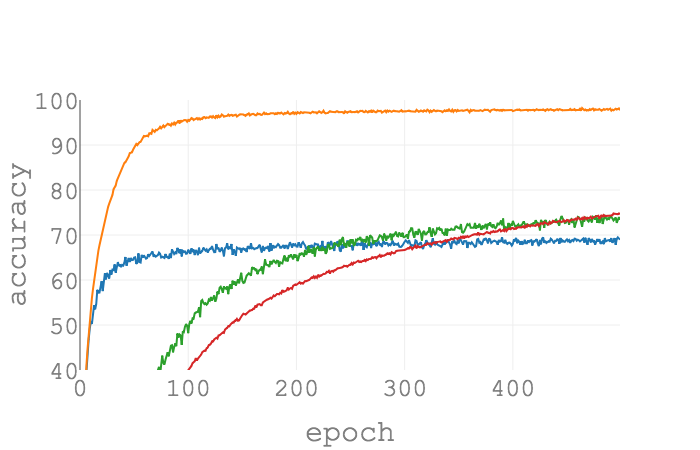}
  \caption{CIFAR-100}
\end{subfigure}
  \begin{subfigure}{.4\textwidth}
  \centering
  \includegraphics[width=80mm,height=40mm]{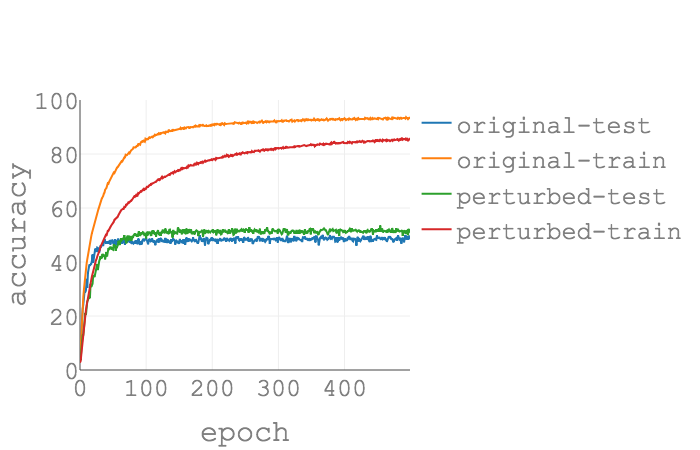}
  \caption{Tiny ImageNet}
\end{subfigure}
\caption{Training and testing accuracy as a function of epochs on CIFAR-10, CIFAR-100 and Tiny ImageNet. For CIFAR, Adam is used as the optimizer, and the learning rate is set as $10^{-4}$. For the Tiny ImageNet, SGD is used as the optimizer, and the learning rate is set as $10^{-2}$.}
\label{fig:pertOPT}
\end{figure}

We compare the perturbed algorithm against the original optimization method on CIFAR-10, CIFAR-100 \citep{CIFAR10}, and Tiny ImageNet \footnote{\url{https://tiny-imagenet.herokuapp.com/}}. The results are shown in Figure \ref{fig:pertOPT}. We use the Wide-ResNet \citep{ZagoruykoK16} as the prediction model.\footnote{\url{https://github.com/meliketoy/wide-resnet.pytorch/blob/master/networks/wide_resnet.py}}
The depth of the chosen model is 58, and the widen-factor is set as 3. The dropout layers are turned off. For CIFAR-10 and CIFAR-100, we use Adam with a learning rate of $10^{-4}$, and the batch size is 128. For the perturbation parameters we use $\eta=0.01$, $\gamma=10$, and $\epsilon$=1e-5. For Tiny ImageNet, we use SGD with learning rate $10^{-2}$, and the batch size is 156. For the perturbed SGD we set $\eta=100$, $\gamma=1$, and $\epsilon$=1e-5. Also we use the validation set as the test set for the Tiny ImageNet. We observe the the effect with perturbation appears similar to regularization. With the perturbation, the accuracy on the training set tends to decrease, but the test or the validation set increases. 

\section{Conclusion}\label{sec:conclud}

We connect the smoothness of the solution with the model generalization in the PAC-Bayes framework. We prove that the generalization power of a model is related to the Hessian and the smoothness of the solution, the scales of the parameters, as well as the number of training samples. In particular, we prove that the best perturbation level scales roughly as $\frac{1}{\sqrt{\nabla^2 \hat{L} + \rho\sqrt{m}\kappa_{\gamma, \epsilon}(w_i)}}$, which mostly cancels out scaling effect in the re-parameterization suggested by \citep{Dinh2017}. To the best of our knowledge, this is the first work that integrate Hessian with the model generalization rigorously, and is also the first work explaining the effect of re-parameterization over the generalization rigorously.  Based on our generalization bound, we propose a new metric to test the model generalization and a new perturbation algorithm that adjusts the perturbation levels according to the Hessian. Finally, we empirically demonstrate the effect of our algorithm is similar to a regularizer in its ability to attain better performance on unseen data. 

\section{Acknowledgement}

The authors are grateful to Tengyu Ma, James Bradbury, Yingbo Zhou, and Bryan McCann for their helpful comments and suggestions on the manuscript.

\vskip 0.2in
\bibliography{library}

\newpage

\appendix

\section{Proof of Lemma \ref{lemma:uniform-pertb-bound}}

\begin{proof}
We rewrite the inequality (\ref{eqn:uniform-bound}) below
\begin{align}
\mathbb{E}_u [L(w+u)]\leq \hat{L}(w) +\frac{1}{6} \sum_i \nabla_i^2 L(w)\sigma_i^2 + \frac{\rho m^{1/2}}{18}\sum_i (\gamma|w_i|+\epsilon)\sigma_i^2 + \frac{\sum_i\log \frac{\tau_i}{\sigma_i} + \log \frac{1}{\delta}}{\eta} + \frac{\eta}{2n} \label{eqn:app-1}
\end{align}

The terms related to $\sigma_i$ on the right hand side of (\ref{eqn:app-1}) are
\begin{align}
\frac{1}{6} \nabla_i^2 L(w)\sigma_i^2 + \frac{\rho m^{1/2}}{18}(\gamma|w_i|+\epsilon)\sigma_i^2 - \frac{\log \sigma_i}{\eta}
\end{align}

Since the assumption is $\nabla^2_i\hat{L}(w^\ast)\geq 0$ for all $i$, $\nabla^2_i \hat{L}(w) + \rho m^{1/2}(\gamma|w_i|+\epsilon)/3> 0$. Solving for $\sigma$ that minimizes the right hand side of (\ref{eqn:app-1}), and we have
\begin{align}
\sigma_i^\ast(w,\eta, \gamma) =\min\left(\sqrt{\frac{1}{\eta(\nabla^2_i \hat{L}(w)/3 + \rho m^{1/2}(\gamma|w_i|+\epsilon)/9)}}, \gamma|w_i|+\epsilon\right)\label{eqn:app-uniform-sigma}
\end{align}

The term 
$\frac{1}{6} \sum_i \nabla_i^2 L(w)\sigma_i^2 + \frac{\rho m^{1/2}}{18}\sum_i (\gamma|w_i|+\epsilon)\sigma_i^2$ on the right hand side of (\ref{eqn:uniform-bound}) is monotonically increasing w.r.t. $\sigma^2$, so
\begin{align}
&\frac{1}{6} \sum_i \nabla_i^2 L(w)\sigma_i^{\ast2} + \frac{\rho m^{1/2}}{18}\sum_i (\gamma|w_i|+\epsilon)\sigma_i^{\ast2} \nonumber\\
 &\leq\sum_i \left( \frac{1}{6}\nabla_i^2 L(w) + \frac{\rho m^{1/2}}{18}(\gamma|w_i|+\epsilon)\right)\frac{1}{\eta(\nabla^2_i\hat{L}(w)/3 + \rho m^{1/2}(\gamma |w_i|+\epsilon)/9)}\nonumber\\
 &=\frac{m}{2\eta}\label{eqn:app-2}
\end{align}

Combine the inequality (\ref{eqn:app-2}), and the equation (\ref{eqn:app-uniform-sigma}) with (\ref{eqn:app-1}), and we complete the proof.

\end{proof}

\section{Proof of Theorem \ref{theorem:uniform-pertb-bound-refine}}

\begin{proof}

Combining (\ref{eqn:app-uniform-sigma-2}) and (\ref{eqn:uniform-bound}), we get
\begin{align}
\mathbb{E}_u [L(w+u)]\leq \hat{L}(w) + \frac{1}{2}\sqrt{\frac{m}{n}} + \frac{\sum_i\log \frac{\tau_i}{\sigma_i^\ast} + \log \frac{1}{\delta}}{\eta} + \frac{\eta}{2n}\nonumber
\end{align}

The following proof is similar to the proof of Theorem 6 in \citep{Seldin2011}. Note the $\eta$ in Lemma (\ref{lemma:uniform-pertb-bound}) cannot depend on the data. In order to optimize $\eta$ we need to build a grid of the form 
\begin{align}
\eta_j = e^j\sqrt{2n\log\frac{1}{\delta_j}}\nonumber
\end{align}
for $j\geq 0$.

For a given value of $\sum_i \log\frac{\tau_i}{\sigma_i^\ast}$, we pick $\eta_j$, such that
\begin{align}
j = \left\lfloor \frac{1}{2}\log\left(\frac{\sum_i\log \frac{\tau_i}{\sigma_i^\ast}}{\log\frac{1}{\delta_j}}+1\right) \right\rfloor\nonumber
\end{align}
where $\lfloor x\rfloor$ is the largest integer value smaller than $x$. Set $\delta_j = \delta 2^{-(j+1)}$, and take a weighted union bound over $\eta_j$-s with weights $2^{-(j+1)}$, and we have with probability at least $1-\delta$, 
\begin{align}
\mathbb{E}_u [L(w+u)]\leq \hat{L}(w) + \frac{1}{2}\sqrt{\frac{m}{n}} + (1+1/e)\sqrt{\frac{\sum_i\log \frac{\tau_i}{\sigma_i^\ast} + \log \frac{1}{\delta} +  \frac{\log 2}{2}\left(2+\log\left(\frac{\sum_i\log \frac{\tau_i}{\sigma_i^\ast}}{\log\frac{1}{\delta}}+1\right)\right)}{2n}}\nonumber
\end{align}

Simplify the right hand side and we complete the proof.

\end{proof}

\section{Proof of Lemma \ref{lemma:gauss-pertb-bound}}

\begin{proof}

We first rewrite the inequality (\ref{eqn:gauss-bound}) below:
\begin{align}
\mathbb{E}_u [L(w+\tilde{u})]\leq \hat{L}(w) + &\frac{1}{2} \sum_i \nabla_i^2 L(w)\sigma_i^2 + \frac{\rho m^{1/2}}{6} \sum_i (\gamma|w_i|+\epsilon)\sigma_i^2 \nonumber\\&+\frac{m\log \tau -\sum_i \log \sigma_i^2 - m + \frac{1}{\tau}\sum_i \sigma_i^2 + 1+ 2\log\frac{1}{\delta} }{2\eta} + \frac{\eta}{2n}\nonumber
\end{align}

The terms related to $\sigma_i$ on the right hand side of (\ref{eqn:gauss-bound}) is 
\begin{align}
\left(\frac{1}{2} \nabla_i^2 L(w) + \frac{\rho m^{1/2}}{6} (\gamma|w_i|+\epsilon)+ \frac{1}{2\tau\eta}\right)\sigma_i^2 - \frac{\log \sigma_i^2}{2\eta}\label{eqn:app-gauss-1}
\end{align}

Take gradients w.r.t. $\sigma_i$, when $\nabla_i^2 \hat{L}\geq 0$, we get the optimal $\sigma_i^\ast$,
 \begin{align}
\sigma_i^\ast = \min\left(\sqrt{\frac{1}{\eta\nabla_i^2 \hat{L}(w^\ast) + \frac{\rho\eta m^{1/2}}{3}(\gamma|w_i^\ast|+\epsilon) + \frac{1}{\tau}}}, \frac{\gamma|w_i^\ast|+\epsilon}{\sqrt{2}\mathrm{erf}^{-1}(\frac{1}{2m})}\right)\nonumber
\end{align}

Note the first term in (\ref{eqn:app-gauss-1}) is monotonously increasing w.r.t. $\sigma_i$, so
\begin{align}
&\left(\frac{1}{2} \nabla_i^2 L(w) + \frac{\rho m^{1/2}}{6} (\gamma|w_i|+\epsilon)+ \frac{1}{2\tau\eta}\right)\sigma_i^{\ast 2}\nonumber\\
&\leq \left(\frac{1}{2} \nabla_i^2 L(w) + \frac{\rho m^{1/2}}{6} (\gamma|w_i|+\epsilon)+ \frac{1}{2\tau\eta}\right)\frac{1}{\eta\nabla_i^2 \hat{L}(w^\ast) + \frac{\rho\eta m^{1/2}}{3}(\gamma|w_i^\ast|+\epsilon) + \frac{1}{\tau}}\nonumber\\
&=\frac{1}{2\eta}
\end{align}

Summing over $m$ parameters and combine (\ref{eqn:gauss-bound}), we complete the proof.

\end{proof}

\section{A Lemma about Eigenvalues of Hessian and Generalization}

\begin{lemma}\label{lemma:app-1}
Suppose the loss function $l(f, x,y)\in [0,1]$. Let $\pi$ be any distribution on the parameters that is independent from the data.  For any $\delta>0$ and $\eta>0$, with probability at least $1-\delta$ over the draw of $n$ samples, for any local optimal $w^\ast$ such that $\nabla \hat{L}(w^\ast)=0$, $\hat{L}(w)$ satisfies the local $\rho$-Hessian Lipschitz condition in $Neigh_{\gamma, \epsilon}(w^\ast)$, and any random perturbation $u$, s.t., $|u_i|\leq \gamma |w_i^\ast|+\epsilon~~\forall i$, we have
\begin{align}
\mathbb{E}_u [L(w^\ast+u)]\leq \hat{L}(w^\ast) +\frac{1}{2} \lambda_{max}\left(\nabla^2 \hat{L}(w^\ast)\right)&\sum_i\mathbb{E} [u_i^2] + \frac{\rho}{6}\mathbb{E}[\|u\|^3] \nonumber\\
&+ \frac{KL(w^\ast+u||\pi) + \log \frac{1}{\delta}}{\eta} + \frac{\eta}{2n}.\label{eqn:lemma-app-1}
\end{align}
\end{lemma}

\begin{proof}
The proof of the Lemma \ref{lemma:app-1} is straight-forward. Since $\nabla \hat{L}(w^\ast)=0$, the first order term is zero at the local optimal point even if $\mathbb{E}[u]\neq 0$. By extrema of the Rayleigh quotient, the quadratic term on the right hand side of inequality (\ref{eqn:lsmooth}) is further bounded by
\begin{align}
u^T \nabla^2 \hat{L}(w) u\leq \lambda_{max}\left(\nabla^2 \hat{L}(w)\right) \|u\|^2. 
\end{align}

Due to the linearity of the expected value, 
\begin{align}
\mathbb{E} [u^T \nabla^2 \hat{L}(w) u]\leq \lambda_{max}\left(\nabla^2 \hat{L}(w)\right) \sum_i \mathbb{E}[u_i^2], 
\end{align}
which does not assume independence among the perturbations $u_i$ and $u_j$ for $i\neq j$.

\end{proof}

\end{document}